\documentclass[letterpaper, 10 pt, conference]{ieeeconf}  % Comment this line out if you need a4paper

\IEEEoverridecommandlockouts                              % This command is only needed if 
\usepackage{graphicx}
\usepackage{caption}
\usepackage{comment}
\usepackage{algorithm}
\usepackage{algorithmicx}
\usepackage{algpseudocode}
\usepackage{amsmath}
\usepackage{color,soul}
\title{\LARGE \bf
A Mini-Review on Mobile Manipulators with Variable Autonomy
}

\author{Cesar Alan Contreras$^{1}$, Alireza Rastegarpanah$^{1, 2}$, Rustam Stolkin$^{1, 2}$, and Manolis Chiou$^{3}$ % <-this % stops a space
\thanks{1 Department of Metallurgy \& Materials Science, University of Birmingham, Birmingham, B15 2TT.}
\thanks{2 The Faraday Institution, Quad One, Harwell Science and Innovation Campus, Didcot, OX11 0RA, UK.}
\thanks{3 Queen Mary University of London, School of Electronic Engineering and Computer Science, London E1 4NS}
\thanks{This work was funded by the Nuclear Decommissioning Authority (NDA) and supported by the National Nuclear Laboratory (NNL).}
\thanks{* Corresponding author: Cesar Alan Contreras, Email: cac214@bham.ac.uk}
}% <-this % stops a space

\begin{document}
\maketitle
\thispagestyle{empty}
\pagestyle{empty}

%%%%%%%%%%%%%%%%%%%%%%%%%%%%%%%%%%%%%%%%%%%%%%%%%%%%%%%%%%%%%%%%%%%%%%%%%%%%%%%%
\begin{abstract}
This paper presents a mini-review of the current state of research in mobile manipulators with variable levels of autonomy, emphasizing their associated challenges and application environments. The need for mobile manipulators in different environments is evident due to the unique challenges and risks each presents. Many systems deployed in these environments are not fully autonomous, requiring human-robot teaming to ensure safe and reliable operations under uncertainties. Through this analysis, we identify gaps and challenges in the literature on Variable Autonomy, including cognitive workload and communication delays, and propose future directions, including whole-body Variable Autonomy for mobile manipulators, virtual reality frameworks, and large language models to reduce operators' complexity and cognitive load in some challenging and uncertain scenarios.
\end{abstract}

\begin{keywords}
Mobile Manipulators, Uncertain Environments, Variable Autonomy, Human-Robot Teaming.
\end{keywords}

%%%%%%%%%%%%%%%%%%%%%%%%%%%%%%%%%%%%%%%%%%%%%%%%%%%%%%%%%%%%%%%%%%%%%%%%%%%%%%%%
\section{Introduction} \label{Introduction}
Robots are being deployed in different environments to aid and complement humans in specific tasks, including manufacturing \cite{rajendran_strategies_2021}, healthcare \cite{lin_shared_2020}, and agriculture \cite{chen_path_2022}, where the benefits of automation are easily observable. Robots have also been used in even more challenging scenarios. For example, mobile robots and mobile manipulators deployed in disaster zones \cite{chen_detection_2019} or in other extreme environments such as nuclear disaster response or decommissioning \cite{nagatani_emergency_2013, chiou_robot-assisted_2022, cragg_application_2003} excel because of their mobility and manipulation capabilities. However, despite their potential, uncertainties prevent these systems from being fully autonomous. Human intervention remains essential due to a lack of trust and the limitations of current autonomous capabilities.

The deployment of autonomous robots across environments shows that no single solution fits all needs. Autonomous systems are limited by their prior knowledge and adaptive capabilities, with training being difficult and time-consuming. Learning from demonstration approaches \cite{moridian_learning_2018} are often confined to their training environments and require human intervention for decisions beyond their training. Improvements in path planning for mobile bases and manipulators \cite{rajendran_strategies_2021, hargas_mobile_2016, chen_path_2022} also face limitations needing human intelligence. Fully manual operation requires significant training and concentration, with cognitive demands varying between operators. These cognitive challenges make fully manual systems more error-prone than semi-autonomous or fully autonomous systems \cite{rastegarpanah_semi-autonomous_2024, chiou_towards_2015}. 

Robots deployed in complex environments increase operator demands for alertness \cite{chiou_flexible_2017}, adaptability to new information \cite{rastegarpanah_semi-autonomous_2024}, and concentration due to unexpected delays and drops in connection \cite{cragg_application_2003}. These burdens can cause physical fatigue and monotony. Some challenges can be alleviated by balancing teleoperation and autonomy, as proposed by field exercises \cite{chiou_robot-assisted_2022}, and reinforced by our anecdotal interactions with operators in Japan preparing for a teleoperated deployment at Fukushima Daiichi Nuclear Power Plant. The operators noted that alertness, concentration, and adaptability are needed due to environmental uncertainties and also due to issues with repetitiveness and communication delays. Automating simple, repetitive processes can address some issues, while other conditions still require manual control. Given the challenges for fully autonomous systems and fully manual systems, developing systems that can switch between human control and autonomy is logical. Mobile bases give the freedom to explore environments, while manipulators enable interaction with objects within them, making mobile manipulators a system worth studying within this context.

% related surveys, and how they have been used in the area
Previous reviews on Variable Autonomy (VA) have focused on cognitive aspects, methodologies, and applications, but not specifically on mobile manipulators. For instance, Tabrez et al. \cite{tabrez_survey_2020} examined mental models and their traits like fluency, adaptability, and effective communication. Villani et al. \cite{villani_survey_2018} discuss cognitive and physical aspects of programming collaborative and shared control robots in industrial settings, emphasizing safe interaction and intuitive interfaces. Bengston et al. \cite{bengtson_review_2020} focused on computer vision for semi-autonomous control of assistive robots, while Moniruzzaman et al. \cite{moniruzzaman_teleoperation_2022} focus on the teleoperation of mobile robots. On the other hand, reviews on mobile manipulators have focused on motion planning \cite{sandakalum2022motion} and the decision-making process of planning algorithms \cite{thakar2023survey} with limited coverage of human-robot interaction, and Variable Autonomy.

Our previous work has addressed varying levels of autonomy in disaster and rescue scenarios, focusing on cognitive and robotic challenges within this scope, limited to mobile robots \cite{chiou_flexible_2017, chiou_mixed-initiative_2021, panagopoulos_hierarchical_2022, ramesh_experimental_2023}. However, there is a need to expand this understanding to other environments where human-robot teams are deployed and mobile manipulators are used. Mobile manipulators can function as single-entity systems, where locomotion and manipulation are coupled, or as dual-entity systems, treating the base and manipulator separately.
With this mini-review, we aim to 1) present the current state of research, 2) identify some challenges, insights, and gaps from the current literature, and 3) propose future research directions with a focus on mobile manipulators, their control within human-robot teams, and varying levels of autonomy.

\section{Methodology} \label{Methodology}
For the review, we performed a search in Google Scholar, with the specifics of our search criteria found in Table \ref{table:search_criteria}. We define mobile manipulators as robots with locomotion decoupled from manipulation, primarily moving on the ground. This definition includes humanoid robots, quadruped or legged robots, wheeled manipulators, and robots on tracks. For the manipulation system, we consider any robot capable of performing tasks typically done by a hand, such as throwing, pushing, grasping, cutting, etc. The system does not necessarily need to perform any manipulation or mobile base task, but it must be physically capable of doing so.

This definition excludes robots with integrated or inseparable locomotion and manipulation systems, such as snake robots, octopus-inspired robots, and soft robots. It also excludes aerial and underwater robots, as well as stationary robotic arms without a mobile base.

\begin{table}[h!]
    \footnotesize
    \centering
    \caption{Search and Exclusion Criteria for the mini-review on Mobile Manipulators with Variable Autonomy}
    \begin{tabular}{|p{6.2cm}|p{1.5cm}|}
    \hline
    \multicolumn{2}{|c|}{\textbf{Search Criteria}} \\
    \hline
    \textbf{Search terms} & \textbf{Results} \\
    \hline
    "mobile manipulator" AND "Variable Autonomy" & 24 \\
    "mobile manipulator" AND "autonomy levels" & 41 \\
    "mobile manipulator" AND "shared autonomy" & 235 \\
    \hline
    \end{tabular}
    
    \vspace{0.2cm}
    
    \begin{tabular}{|p{1.5cm}|p{6.2cm}|}
    \hline
    \textbf{Period} & January 2018 – June 2024 \\
    \hline
    \textbf{Search engines} & Google Scholar \\
    \hline
    \textbf{Publication type} & Peer-reviewed papers, academic papers, conference papers, pre-prints, journal articles, theses.\\
    \hline
    \end{tabular}
    
    \vspace{0.0cm}
    
    \begin{tabular}{|p{1.5cm}|p{6.2cm}|}
    \hline
    \multicolumn{2}{|c|}{\textbf{Exclusion Criteria}} \\
    \hline
    \textbf{Language} & Non-English \\
    \hline
    \textbf{Contextual} & Studies involving robots which aren't contained in the mobile manipulators (e.g., aerial vehicles, underwater robots, soft robots) \\
    \hline
    \textbf{Scope} & Excluded papers not addressing or utilizing advancements in Variable Autonomy(e.g mixed initiative, shared control, etc) \\
    \hline
    \end{tabular}
    
    \label{table:search_criteria}
\end{table}

Based on the number of search results, most research is centered around shared autonomy, rather than mixed initiative or other forms of Variable Autonomy. After applying the exclusion and contextual criteria, \textbf{38} papers were included in the review.

\section{Literature Review}  \label{Review}
Variable Autonomy systems enable flexible control by humans and machines across different levels of operation. Although various definitions exist for Variable Autonomy in the literature \cite{Reinmund_2024_Variable, Methnani_2024_Who}, we classify these systems into two primary categories for this review. The first is autonomy-changing systems, where autonomy levels can be modified during task execution. Full VA or Mixed-Initiative systems allow humans and machines to adjust these levels. Only humans can make changes in human-initiative (HI) or adjustable autonomy systems, whereas AI-initiative (AI-I) or sliding autonomy systems permit machines to do so. The second category is autonomy-sharing systems, often called Shared Control systems. In these systems, humans and robots work together on tasks. Shared Control can be either supervisory, where humans provide high-level directives, or assistive, where humans directly control the robot with system support, such as visual feedback or trajectory guidance. 

\subsection{\textbf{Environments}}
The first focus is on the deployment environments. For this categorization, we examined the main ideas and purpose of each paper, as long as the application environment is detailed in their methodology. This condition identified seven primary environments: hazardous materials and environments handling, disaster response, industrial applications and manufacturing, Research and Development (R\&D) in laboratories, healthcare and medical applications, agriculture or farming, and domestic and household environments. Table \ref{table:paper_summary} lists the papers by environment and provides brief descriptions and applications.

\begin{table*}[h!]
    \scriptsize
    \centering
    \caption{Summary of Papers, Environments, and Key Findings}
    \begin{tabular}{|p{2.5cm}|p{2.5cm}|p{11cm}|}
    \hline
    \textbf{Papers} & \textbf{Environment} & \textbf{Summary} \\
    \hline
    \cite{frese_autonomous_2022, roennau_grasping_2022, schuster_arches_2020, wedler_preliminary_2021, woock_robdekon_2022} & Hazardous materials handling & Discuss the use of mobile manipulators in handling hazardous materials, examples include digging contaminated soil or waste \cite{frese_autonomous_2022, woock_robdekon_2022}, grasping hazardous objects \cite{roennau_grasping_2022}, and taking space soil samples \cite{schuster_arches_2020, wedler_preliminary_2021}.
    \\
    \hline
    \cite{chiou_robot-assisted_2022, frering_enabling_2022, verhagen_meaningful_2024} & Disaster response & Applications examples include the deployment of mobile manipulators for exploration and mapping of destroyed nuclear facilities \cite{chiou_robot-assisted_2022} and scenarios for fire fighting \cite{frering_enabling_2022, verhagen_meaningful_2024}. \\ 
    \hline
    \cite{merkt_towards_2019, sirintuna_enhancing_2024, stibinger_mobile_2021} & Industrial manufacturing & Examine applications in industrial manufacturing, examples include performing tasks on moving assembly components \cite{merkt_towards_2019}, collaborative transportation of large objects \cite{sirintuna_enhancing_2024} and placement of construction materials \cite{stibinger_mobile_2021}. \\
    \hline
    \cite{moridian_learning_2018, hargas_mobile_2016, baberg_improving_nodate, baek_study_2022, benzi_whole-body_2022, chen_humanrobot_2018, cheong_supervised_2021, fozilov_towards_2021, gholami_shared-autonomy_2020, li_automatic-switching-based_2024, palan_learning_2019, valner_example_2018, wong_touch_2022} & Research and development laboratories & Reviews the use of mobile manipulators in R\&D labs, focusing on research and experiments not for a specific deployment. Examples include works on teleoperation \cite{baberg_improving_nodate, chen_humanrobot_2018, cheong_supervised_2021, gholami_shared-autonomy_2020, li_automatic-switching-based_2024, valner_example_2018}, path planning \cite{moridian_learning_2018, baek_study_2022, hargas_mobile_2016, palan_learning_2019}, whole-body control \cite{benzi_whole-body_2022, wong_touch_2022}, and multi-robot coordination \cite{fozilov_towards_2021} \\
    \hline
    \cite{lin_shared_2020, abubakar_arna_2020, kapusta_task-centric_2019, sanchez_shared_2021, sanchez_verifiable_2022} & Healthcare and medical areas & Investigates applications in healthcare, including nursing, patient care \cite{abubakar_arna_2020, kapusta_task-centric_2019, lin_shared_2020} and surface disinfection \cite{sanchez_shared_2021, sanchez_verifiable_2022}. \\
    \hline
    \cite{chen_path_2022} & Agriculture and farming & Studies the use in agricultural settings, in this case, the example applies to the pruning of a tree \cite{chen_path_2022}. \\
    \hline
    \cite{bhattacharjee_is_2020,  karim_investigating_2023, kemp_design_2022, kim_dynacon_2023, mirjalili_lan-grasp_nodate, park_active_2020, pohl_makeable_2024, rakita_shared_2019} & Domestic and household environments & Includes the use of mobile manipulators in the house, examples include feeding assistance \cite{bhattacharjee_is_2020, karim_investigating_2023, park_active_2020}, and general house grasping and transportation tasks \cite{kemp_design_2022, kim_dynacon_2023, mirjalili_lan-grasp_nodate, pohl_makeable_2024, rakita_shared_2019}. \\
    \hline
    \end{tabular}
    \label{table:paper_summary}
\end{table*}

\subsection{\textbf{Tasks that Variable Autonomy tackles}}

After categorizing the papers by environment, the next step is to categorize them by the specific tasks to which Variable Autonomy is applied. This categorization means focusing on how Variable Autonomy and human-robot interaction are utilized as tools to accomplish various tasks. In other words, while some papers include the use of Variable Autonomy, they do so to aid in completing other tasks, and not necessarily researching ways to change the autonomy levels.

\textbf{Human Mapping movement.} Humanoid robots often take inspiration from human capabilities \cite{lin_shared_2020, baek_study_2022, pohl_makeable_2024}. Some of these systems use motion mapping based on human posture or control. For instance, Baek et al. \cite{baek_study_2022} utilize human leaning, to control velocity while avoiding obstacles. In this approach the robot provides a force feedback based on proximity to obstacles, allowing the user to adjust their input. The robot can also alter its own velocity and path. Similarly, Lin et al. \cite{lin_shared_2020} map human movements to specific humanoid manipulations to simplify a pick-and-place task. In their system, the human only needs to make the physical movement sign, and the robot autonomously completes the task after the decision has been made.

\textbf{Manipulation.} Research in manipulation includes grasping, autonomous manipulation, and load balancing. This type of research is characterized by robots performing tasks with some level of autonomy. Papers that involve manipulation research in any form include \cite{lin_shared_2020, chen_path_2022, hargas_mobile_2016, frese_autonomous_2022, roennau_grasping_2022, schuster_arches_2020, wedler_preliminary_2021, merkt_towards_2019, stibinger_mobile_2021, cheong_supervised_2021, fozilov_towards_2021, kapusta_task-centric_2019, sanchez_shared_2021, sanchez_verifiable_2022, mirjalili_lan-grasp_nodate, park_active_2020, pohl_makeable_2024, rakita_shared_2019}
. Unlike teleoperation research, the robot completes most of the manipulation tasks by itself in this area. For example, Frese et al. \cite{frese_autonomous_2022} use a giant excavator that plans how to dig soil, understands its properties, and balances the load. The system allows it to ask for the help of an operator in case the probability of success decreases. Another example is Cheong et al. \cite{cheong_supervised_2021}, who gives the operator the option to select the object he wants to manipulate, but the robot extracts key points and finds grasping candidates to grab the object by itself.

\textbf{Transportation.} This area includes path planning that considers the geometry or other physical properties of the objects being moved, as well as obstacles and humans in the path of the robot or object \cite{woock_robdekon_2022,  sirintuna_enhancing_2024, benzi_whole-body_2022, abubakar_arna_2020}. For example, ARNA, from Abubakar et al. \cite{abubakar_arna_2020} can transport an object while assisting a patient walking through a scene. In this case, the patient only controls the direction, while the robot independently manipulates and transports the object. Sirintuna et al. \cite{sirintuna_enhancing_2024} proposed a collaborative approach where the robot provides haptic feedback through a belt worn by a human in an occluded environment. Assisting them in transporting an object collaboratively with the robot by feeling a force when obstacles get closer. The human commands the direction, while the robot provides environmental information and transports the vehicle with a fixed end effector position relative to the base.

\subsection{\textbf{Challenges and techniques VA aids with.}}
\textbf{Collision, Obstacle Avoidance, Mapping and Navigation.} This area utilizes sensor integration and real-time processing to enable robots to make decisions and adjust their path to avoid collisions. Relevant papers include \cite{hargas_mobile_2016, frese_autonomous_2022, roennau_grasping_2022, woock_robdekon_2022, cheong_supervised_2021, fozilov_towards_2021, gholami_shared-autonomy_2020, valner_example_2018, kapusta_task-centric_2019}. Roennau et al. \cite{roennau_grasping_2022} describe a system where an operator selects an object to retrieve, and the robot plans the path while using a 3D SLAM-generated map it to avoid any possible collisions. Another example is Valner et al. \cite{valner_example_2018}. They developed a framework where the machine autonomously changes a sensor feed to another if a redundant sensor fails. The human operator provides high-level commands, asking the robot to capture the environment while the system handles the mapping details.

% baberg worked in an interface that tells the operator where the network is going to be less reliable
\textbf{Communication and Delays.} The impact of communication delays is discussed by various researchers \cite{ frese_autonomous_2022, merkt_towards_2019, baberg_improving_nodate, li_automatic-switching-based_2024, valner_example_2018}. Baberg et al. \cite{baberg_improving_nodate} have shown some work in user interface that helps an operator assess network reliability. Frese et al. \cite{frese_autonomous_2022} depend on hardware communication speeds with buffer configuration and pre-allocation of memory. Variable Autonomy can help mitigate delays by providing autonomous control when high latency is detected, running directly on the robot's internal systems, while allowing long-distance manual control when latency is low. 

\textbf{Semantics and machine learning.} This area takes advantage of the computational power for object recognition \cite{lin_shared_2020, woock_robdekon_2022, cheong_supervised_2021, bhattacharjee_is_2020, park_active_2020, pohl_makeable_2024}, task learning \cite{wong_touch_2022, park_active_2020, rakita_shared_2019}, and the use of Large Language Models (LLMs) \cite{kim_dynacon_2023, mirjalili_lan-grasp_nodate} for developing smarter systems. The main focus of the area is helping a human reduce their cognitive load, smarter systems can allow the human to take a supervisory role in the tasks, and only take full control when an object or task not previously trained for is encountered. As an example Bhattacharjee et al. \cite{bhattacharjee_is_2020} allow a user to select a food from an interface, limiting its choices to some fruits detected by a perception algorithm but allowing the user to take manual control of other feeding processes.

\textbf{Teleoperation Modes, and intent recognition.} Popular research areas that allow humans to manually control a robotic system from a distance or share the control with the robot \cite{schuster_arches_2020, verhagen_meaningful_2024, baek_study_2022, chen_humanrobot_2018, gholami_shared-autonomy_2020, li_automatic-switching-based_2024, wong_touch_2022, bhattacharjee_is_2020, kemp_design_2022}. Some researchers focus on providing high-level commands, enabling the robot to execute pre-programmed tasks while they explore other methods to communicate their intentions. Bhattacharjee et al. \cite{bhattacharjee_is_2020} utilize voice commands, Wong et al. \cite{wong_touch_2022} try influencing a robot with physical touch, and Chen et al. \cite{chen_humanrobot_2018} propose utilising hand gestures. Other researchers use computer assistance for specific tasks, while manually moving the robots. Li et al. \cite{li_automatic-switching-based_2024} manage teleoperation of the mobile base and manipulator arm independently, but use the system to decide when to switch between devices. Baek et al. \cite{baek_study_2022} implement full-body teleoperation in VR, relying on the human inclination to adjust the robot's speed and movement while the computer assists him in avoiding collisions.

\section{Insights and challenges} \label{Discussion}

The explored literature on Variable Autonomy for mobile manipulators is divided into two focuses: \textbf{1)} high-level control, or supervisory control, and \textbf{2)} low-level control or system assistance.  Most implementations involving manipulation, obstacle avoidance, mapping, transportation, and machine learning research aim for fully automated tasks. In these cases, the role of the operator is primarily to decisions, choose tasks, or supervise to ensure the robot is not making mistakes. For known problems, this solution is good, providing automated solutions that are becoming easier for non-expert humans to use. On the other hand, we have teleoperated scenarios often with some uncertainty. In these, humans drive the base or move the arm, with autonomy serving in an assistive capacity. The main objective here is to lower human cognitive load or to reduce the operation completion time.

\textbf{A) Separate Focus on Base and Manipulator} In mobile manipulators, the use of varying levels of autonomy is still primarily focused on either the base or the manipulator separately. Current research does not consider the joint problem of integrating changes in autonomy for both. This can be seen in a multitude of papers including: \cite{lin_shared_2020, frese_autonomous_2022, woock_robdekon_2022, merkt_towards_2019, baberg_improving_nodate,  cheong_supervised_2021, fozilov_towards_2021,  gholami_shared-autonomy_2020, palan_learning_2019, valner_example_2018, sanchez_verifiable_2022,  bhattacharjee_is_2020, karim_investigating_2023, kemp_design_2022,   kim_dynacon_2023, mirjalili_lan-grasp_nodate, park_active_2020, pohl_makeable_2024, rakita_shared_2019}. Researchers in this area focus on applying varying levels of autonomy to either the base or the manipulator, while keeping the rest of the robot static or at the same autonomy level throughout the task. Even transportation tasks follow a sequential process of changing between both: reaching a position with the base, picking the object with the manipulator, reaching a dropping position with the base, and placing the object with the manipulator. This approach can theoretically limit the operational workspace of a mobile manipulator. For example, in \cite{sanchez_verifiable_2022} the disinfection area is limited because the mobile base is not used simultaneously to increase the reach of the manipulator. 

\textbf{B) Human Cognitive Load.} Refers to the mental effort required to perform a task and is a term acknowledged and investigated in multiple papers including, \cite{chiou_flexible_2017, sirintuna_enhancing_2024, baek_study_2022, lin_shared_2020}. However, it is still primarily studied using subjective measures, such as the NASA Task Load Index (NASA-TLX). Currently, objective metrics and biometric data from the human operator are not widely used in systems with mobile manipulators with variable autonomy or involving human-in-the-loop operations. Implementing objective data from the participants would enhance our understanding of cognitive load and help in designing better support for human operators

\textbf{C) Communication Delays, and System Reliability.} Although known to cause issues, they are often ignored or not measured in implementations of mobile manipulation. There is a lack of studies addressing this problem in relevant environments. Moniruzzaman et al. \cite{moniruzzaman_teleoperation_2022} mention the existence of compensation techniques such as future pose estimation and point-cloud 3D reconstruction, that could benefit the area if applied. In addition, Variable Autonomy could be used by switching from manual teleoperation to a local compensation algorithm when higher latency is detected.

\textbf{D) Uncertain Environments.} Applying varying levels of autonomy in known environments, where high-level and supervisory control is feasible, is a popular and researched area. The challenge lies in extending high-level control strategies to more complex and unpredictable environments, where robust decision-making and adaptability matter.

\section{Future Work}
\vspace{-2mm}
\textbf{1) Better system integration.} Future research should focus on enabling the simultaneous control of both the base and manipulator, whether coupled or decoupled. This would allow to switch levels of autonomy for both. This helps in extending research that already allows the autonomous switching of operator control between the base and manipulator \cite{li_automatic-switching-based_2024}. This would allow operation on larger manipulation workspaces. 

\textbf{2) Virtual Reality (VR).} Some papers already demonstrate the opportunities that VR offers in reducing cognitive load and improving environmental awareness \cite{baek_study_2022, woock_robdekon_2022, rastegarpanah_semi-autonomous_2024}. Future research could explore implementing VR in more challenging mobile manipulation problems, taking into account world physics and virtual world design metrics for a smoother operator experience. 

\textbf{3) Machine Learning for delay or error reduction.} Machine learning techniques, such as intent recognition, are already used in high-level control to assign tasks to robots \cite{gholami_shared-autonomy_2020} and to learn reward functions from human demonstrations \cite{palan_learning_2019}. There is an opportunity to use these techniques to continue tasks that generally would be dropped because of latency. Integrating onboard fully autonomous compensation or task algorithms in long-distance systems which can also be manually teleoperated, can help mitigate problems. Using autonomy switching between both can help compensate for the delays and errors that might occur. 

\textbf{4) More LLMs.} Large Language Models, trained with the data available on the net, are becoming multi-modal. There is an opportunity to use these systems as general assistants to numerous tasks. Kim et al. already demonstrate how contextual awareness is possible with this tool \cite{kim_dynacon_2023}. LLMs could enhance task awareness and be used as a dynamic way to adjust autonomy levels based on new information not considered in the original algorithms. This could lead to the development of on-demand algorithms, improving the flexibility of mobile manipulators.
\vspace{-2mm}
\section{Conclusion} \label{Conclusion}
\vspace{-2mm}
This mini-review synthesized current research on mobile manipulators with varying degrees of autonomy, revealing gaps and possible opportunities. The gaps included: First, the autonomy of mobile manipulators often focuses separately on the base and the manipulator. Second, studies on cognitive workload are heavily based on subjective metrics. Third, communication delays and reliability issues are acknowledged but not extensively addressed. Finally, most of the research is environment-specific and lacks implementations in uncertain environments. Future research should aim to develop integrated autonomy for both the base and the manipulator, use VR or other intuitive interfaces, implement adaptive communication protocols to handle network instability, and real-time general decision-making frameworks in real time based on LLMs that dynamically adjust autonomy levels based on situational demands. In addition to this paper, an extensive mobile manipulators literature survey is needed, including advancements on mobile robots and manipulator arms separately, to justify the findings of this mini-review further. 

%%%%%%%%%%%%%%%%%%%%%%%%%%%%%%%%%%%%%%%%%%%%%%%%%%%%%%%%%%%%%%%%%%%%%%%%%%%%%%%%

\bibliographystyle{IEEEtran}
\bibliography{references_new}

\end{document}